\documentclass{article}

\usepackage{arxiv}

\usepackage[utf8]{inputenc} 
\usepackage[T1]{fontenc}    
\usepackage{hyperref}       
\usepackage{url}            
\usepackage{booktabs}       
\usepackage{amsfonts}       
\usepackage{nicefrac}       
\usepackage{microtype}      
\usepackage{lipsum}
\usepackage{microtype}
\usepackage{multicol}
\usepackage{graphicx}
\usepackage{xspace}
\usepackage{multirow}
\usepackage{xcolor}
\usepackage{amsmath}

\usepackage{xspace}
\usepackage{subcaption}
\usepackage{amsthm}

\newcommand{\ourm}{\textsc{RelBert}\xspace}

\newcommand{\cm}[1]{\mathcal{#1}}

\newcommand{\bs}[1]{\boldsymbol{#1}}
\newif\ifdraft
\draftfalse
\ifdraft
    \newcommand{\sbcomment}[1]{\textcolor{red}{{\bf SB:}#1}}
\else
    \newcommand{\sbcomment}[1]{}
\fi

\theoremstyle{definition}
\newtheorem*{problem*}{Problem Statement}

\title{BERT meets Relational DB: Learning Deep Contextual Representations of Relational Databases}

\author{
 Siddhant Arora, Vinayak Gupta, Garima Gaur, Srikanta Bedathur\\
 Indian Institute of Technology, Delhi \\
 New Delhi, India 110016\\
}

\begin{document}
\maketitle

\begin{abstract}
In this paper, we address the problem of learning low dimension representation of entities on relational databases consisting of multiple tables. Embeddings help to capture semantics encoded in the database and can be used in a variety of settings like auto-completion of tables, fully-neural query processing of relational joins queries, seamlessly handling missing values, and more. Current work is restricted to working with just single table, or using pretrained embeddings over an external corpus making them unsuitable for use in real-world databases. In this work, we look into ways of using these attention-based model to learn embeddings for entities in the relational database. We are inspired by BERT style pretraining methods and are interested in observing how they can be extended for representation learning on structured databases. We evaluate our approach of the autocompletion of relational databases and achieve improvement over standard baselines.
\end{abstract}

\section{Introduction}
\label{sec:introduction}

Majority of the structured data on the Web is modeled as tabular data. Many search engines, like Google Search and Microsoft's Bing, rely on semi-structured Wikipedia infoboxes and tables as their primary source of information. Majority of data, specifically industrial data, is stored and managed by good old RDBMS systems like IBM Db2, Microsoft SQL server, Oracle DB, etc. The success of deep learning models \cite{BERT,lstm,GPT} in addressing the NLP tasks has lead to plethora of work \cite{Bordawekar,TableBERT,DeepImputation,turl,TAPER} that try to embed the structured tabular data to a latent space. Good quality embeddings can play crucial role in improving downstream database specific tasks like, \emph{imputation of missing values} \cite{Holoclean}, \emph{data integration} \cite{EmbDi}, \emph{schema matching} \cite{schema-matching}, \emph{cognitive query evaluation} \cite{bordwaker}.

Existing powerful NLP models \cite{GPT,BERT} cannot be used off-the-shelf to learn embedding in relational data. The structural difference between the textual and tabular data makes it hard to use the existing learning models out of the box. Relational databases are tightly coupled with a coherent schema that defines the structure of the data. Therefore, data let alone does not capture the entire semantics of the data stored. 
Conversely, just the table specific information, like column headers, might not be very informative either. For example, a table storing information about people can have columns \texttt{f\_anme} \texttt{l\_name} to store their \emph{first name} and \emph{last name} respectively. Therefore, just by looking at the column name it might not be feasible to understand the information residing in the column. Therefore, non-tweaked existing models, that are trained over huge text corpus, cannot be used in practice to get effective embeddings of entities stored in a database. 

\begin{figure}[t]
    \centering
    \includegraphics[scale= 0.6]{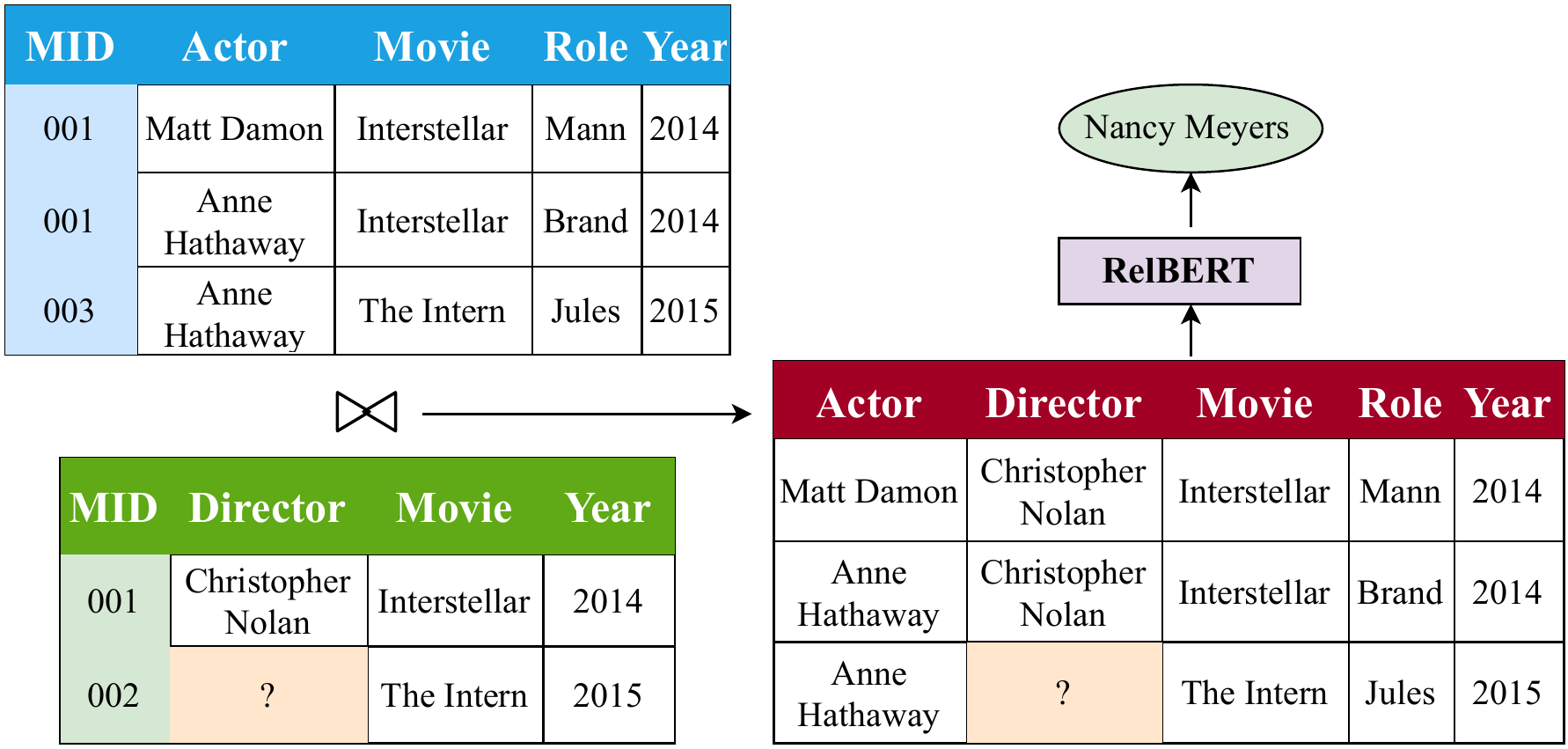}
    
    \caption{Schematic diagram of Auto-Completion procedure of \ourm with two tables from the IMDB dataset}
    \label{fig:intro}
\end{figure}

Some of the recent BERT style models \cite{turl,TableBERT} have worked with web-tables and therefore miss out on the intricacies of a relational database, like domain-specific vocabulary, \emph{normalized} data spread over many tables  and  integrity constraints.
Other models \cite{bordwaker,TAPER,Holoclean,CrimeRel} that deal with the relational data treat rows of a table as standalone sentences and use the corpus of such sentences to train word2Vec model\cite{Word2vec}. The \emph{state-of-the-art} model Embdi \cite{EmbDi} tried a sophisticated approach based on random walks to capture intra-column relations between entities, but they also did not take into account the relationships across tables. These relationships are often expressed as \emph{foreign key}-\emph{primary key} (FK-PK) joins. Often relation databases are \emph{normalized} and therefore relationship among entities might not be directly present in a row of a table. For instance, consider IMDB database that store information about movies, web-series, documentaries, etc. The complete description of a movie would involve information about its director and actors. This information might be stored in two different tables of the database. Therefore, to learn better contextual embeddings it is essential to consider these semantic relationships across tables.

To address these shortcomings of existing approaches, we propose \ourm, an attention based model that learns entity embeddings via  capturing semantic relationships across tables. We define semantics of entities based on the table column they appear in. For instance, \textit{Clint Eastwood} is an actor as well as a director, syntactically the entity \textit{Clint Eastwood} in column \texttt{director} and \texttt{actor} is same but semantically they are different. Therefore, to incorporate relational data aware semantics we map entities of different columns to different latent spaces. 
We propose two task-specific variants: (i) \ourm-A and (ii) \ourm-J. For missing value imputation, we propose a generic model \ourm-A that performs BERT-style \emph{masked language model} training to learn contextual embeddings to entities in the database. \ourm-J, on the other hand, learns to predict joins as  NSP task. We evaluate the performance of our model using two real databases, IMDB \footnote{\url{https://relational.fit.cvut.cz/dataset/IMDb}} and MIMIC \footnote{\url{https://mimic.physionet.org/}} and report an averaged (across all metrics) performance gain  of $11.3\%$ and $41.6\%$ respectively over the state-of-the-art table embedding model EmbDI \cite{EmbDi}.

\section{Related Work}
\label{sec:relatedWork}

Efforts towards mapping tabular data to a latent space  has lead to many works \cite{Bordawekar,BalogTableSearch,CrimeRel,InfoGather,EmbDi,TableBERT}. Each focusing on data at different granularity, ranging from representing the entire table as a single vector \cite{BalogTableSearch,RelatedTable} to mapping each token in a table cell to a vector \cite{EmbDi,bordwaker}. 
Earlier approaches (\cite{BalogTableSearch,RelatedTable}) learnt the table embedding with the intent of finding related tables. Recently, focus shifted to addressing downstream tasks like, missing value prediction, join predictions, cognitive query processing, schema matching, etc, using entity embeddings. 

\cite{Bordawekar,RajeshDisclosure} focused on answering cognitive queries over relational data that involve finding similar, dissimilar or analogues entities. Their approached involved interpreting rows of tables as natural language sentences and then train a word2vec model to embed entities in a latent space. \cite{CrimeRel} aimed to use learned embeddings to identify nontrivial patterns from the database that helps in predicting appropriate policing matters. This way of simply interpreting table rows as NL sentences fails to capture the semantic relations between tables which are often expressed via \emph{foreign key and primary key} (FK-PK) pairs. We leverage such implicit information about data that is encoding in the database schema. 
Efforts in direction of addressing missing value completion problem has lead to  two successful approaches, \cite{DeepImputation} and \cite{Holoclean}. However, due to sole focus of these models on missing value completion task, the learned embeddings are quite task specific and fail to serves as good quality embeddings for other downstream tasks. 

The \emph{state-of-the-art} model EmbDi(\cite{EmbDi}) that learns embeddings of entites stored in a RDBMS  relies on random walks for the sentence generation. They construct a tripartite graph with each entity connected to row ids and column ids in which it resides. Random walks over this graph created sentences that contain entities that might not directly be present in same row or same column in the table. They claim to capture intra-column dependencies with this way of constructing sentences. They also employed the standard NLP-based embedding approach i.e., word2vec, over this sentence corpus to generate embeddings. Some of the recent attention based models \cite{TableBERT,turl} work with web-tables. Therefore, they do not deal with the relational data specific  challenges like domain dependent custom vocabulary (like patient ids, drug codes used in MIMIC), semantic relationships across tables, imposed integrity constraints, etc. 

Most of the approaches either work with web-tables and, thus, do not deal with the intricacies of a relational database or, even if they work with the relational data, they fail to fully leverage the implicit knowledge a database schema captures about the stored data. In this work, we do not treat entities just as a simple word in a NL sentence rather, we focus on their semantics. For instance, if a syntactically same entity is present in two different columns, we treat them as separate entities. Further, unlike previous works, we do not simply treat each row of a table as a sentence. Our approach focused on capturing semantic relationships across tables that are expressed as FK-PK joins between tables in a database.

\section{Problem Description}
We consider a relational database $\bs{\cm{D}}$ consisting of a set of  tables denoted by $\bs{T}_{(\alpha, \beta \cdots)} \in \bs{\cm{D}}$. For each table, $\bs{T}_{\bullet}$, we denote its \textit{attributes}(or columns) as $\bs{T}_{\bullet}^{(A,B \cdots)}$ and each \textit{tuple}(or row) is addressed via its \textit{primary key}. We denote the number of columns in a table as $\cm{C}_{\bs{T}_{\bullet}}$.
Additionally, to include the data spread across different tables, we employ a \textit{joined} version of two tables, say $\bs{T}_{\alpha}$ and $\bs{T}_{\beta}$, with all \emph{joins} materialized via their \textit{primary-key}, $\tau_i \in \bs{T}_{\alpha}$, and the corresponding \textit{foreign-key}, $\tau_j \in \bs{T}_{\beta}$. For an intelligible model description, we slightly abuse the notation scheme and denote the resulting table as $\bs{T}_{\alpha \beta}$ or $\bs{T}$ in general. For simplicity of exposition, we ignore the possibility that after the join step we may have multiple (unrelated) tables. 

Since \ourm deploys a self-attention~\cite{BERT} based encoder, we consider each row in $\bs{T}$ as a \textit{sentence}. For example in Figure ~\ref{fig:intro}, we consider the row \textit{"Matt Damon-Christopher Nolan-Interstellar-Mann-2014"} as an input sentence. Our aim is to utilize these sentences and the relational structure to learn vector representations for each entity in the table. We focus on two principal database-problems of Table \textit{Auto-completion} and \textit{Join Prediction}. 
\begin{problem*}[\textbf{Table Autocompletion}]
\label{problem}
\textit{Following a \textit{masked-language} based learning procedure, we randomly mask one of the entities in a sentence from our model and for our autocompletion problem, our aim via \ourm is to get a ranked list of most probable candidates for the masked entity. Specifically, maximize the function: $\mathbb{E}[\widehat{v} \in \bs{T}_i| \widehat{v} \in \bs{T}^{A}] \, \forall \bs{T}_{i}$, where $\mathbb{E}[\widehat{v}]$ calculates the probability of a candidate entity to be the masked entry in the sentence and $\bs{T}^{A}$ denoted the column of the masked entity.}
\end{problem*}

\begin{problem*}[\textbf{Join Prediction}]
\label{problem}
\textit{Given two join-compatible tables $\bs{T}_{\alpha}$ and $\bs{T}_{\beta}$ that are joined over $\tau_{i}$ ($\in \bs{T}_{\alpha}^{(A,B,\dots)}$) and $\tau_{j}$ ($\in \bs{T}_{\beta}^{(A,B,\dots)}$) --i.e., $\bs{T}_{\alpha} \Join_{\tau_i, \tau_j} \bs{T}_{\beta}$ is well defined-- \ourm predicts tuples of $\bs{T}_{\beta}$ that would join with a given tuple of $T_{\alpha}$, i.e., the given tuple and the predicted tuple are expected to agree on the respective value of $\tau_{i}$ and $\tau_{j}$.}
\end{problem*}

In this paper we propose two variants corresponding to both the database problems. For table autocompletion, we propose \ourm-A, a table-encoder model that captures relationships between tables via a Masked-Language Model (MLM) and a inter-relation Next Sentence Prediction (NSP). For join-prediction problem, we propose \ourm-J that evaluates the learned embeddings by computing the variations between a complete \textit{table-based} join and an embedding based \textit{neural}-join.

\section{Proposed Model}
In this section we first describe in detail the underlying model of \ourm that is consistent with both the variants \ourm-A and \ourm-J. Later, we give a detailed difference between the distinct learning-procedures for both the models. We further describe them here:

\subsection{Table Encoder} 
For any table $\bs{T}_{\bullet}$, our model incorporates a self-attention model for encoding the entities in a table. As a sequential input to the transformer, we use the all the sentences (or rows) in a table. However unlike a natural language sentence, each word in a row belongs to a column in the table and each column inherently contains distinct entity information. For example, in Figure 1 each entity in the sentence corresponds to a distinct attribute of an \textit{actor/director} or the \textit{movie-name}. Thus, a homogeneous embedding initialization function will fail to incorporate the column heterogeneity. Therefore, we use distinct embedding spaces corresponding to each column in the table denoted by $\Phi^{\bullet}$. We calculate the embedding matrix for the entities in the table as 
\begin{equation}
\bs{P}^k_i = \Phi^{k}(\bs{T}^k_i), \qquad \forall k, i \in \bs{T},
\end{equation}
where $\Phi^k$ and $i$ denote embedding function for $k$-th column and the row (or sentence) index in the relational table respectively.
Following ~\cite{BERT}, we utilize a masked-language based learning where we randomly mask one of the entities in a sentence and then predict those masked entries based on the transformer output. We compute the output embeddings for each entity through an encoder network consisting of multiple attention layers. We denote the output embeddings for each sentence indexed by $i$ as $\bs{O}_i = \texttt{TEncoder} (\bs{P}_i)$, where $\texttt{TEncoder}(\bullet)$ denotes the base-transformer module.
\begin{figure}[t]
\centering
\includegraphics[scale=0.7]{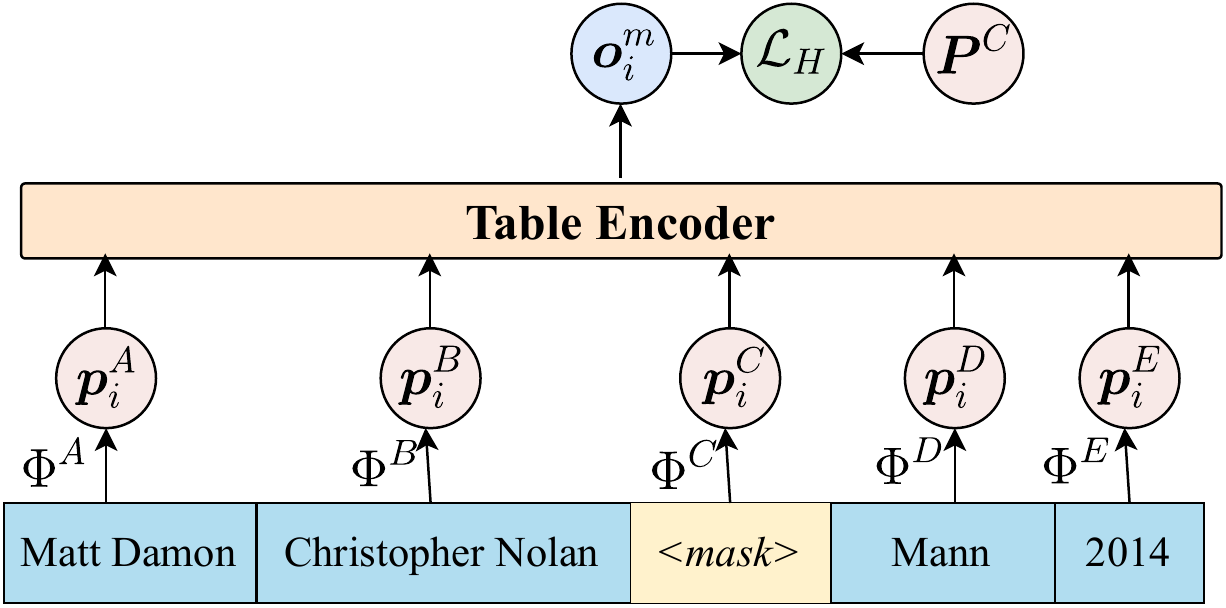}
\caption{Schematic diagram showing the training of the proposed Transformer Encoder based Language Model.}
\label{fig:MLM_Encoder}
\end{figure}
However, since each column $k$ has a different embedding space, computed by $\Phi^k$, we determine the candidate entities for the \textit{masked} entity in the sentence using an output-softmax over the entities in the column of the masked entity. We learn the parameters for the horizontal transformer by optimizing the cross entropy loss function for each sentence in the table $\bs{T}$ as:
\begin{equation}
\cm{L}_{mlm} = \sum_{i=1}^{|\bs{T}|}\texttt{CrossEntropy}(\bs{o}^{m}_i, \bs{P}^{mask}_i), \qquad \bs{o}^m_i \in \bs{O},
\end{equation}
where, $\bs{o}^m_i, \bs{P}^{mask}$ denote the output embedding corresponding to the masked index and the column entities respectively. In addition, we exclude positional embeddings from the transformer model as a table is inherently \textit{permutation-invariant}. We initialize each entity in a row of the table using standard word2vec~\cite{Word2vec} computed over the table by simply treating each row as a sentence (for example as in~\cite{bordwaker}).


\subsection{\ourm-A}
This variant of \ourm learns representations for all entities in order to impute the missing entries in a table. However, a relational table often is organized into multiple tables either to ensure integrity constraints efficiently and to improve the overall database performance using \emph{data normalization} process. Therefore, in order to get a list of candidates for a missing entry in a table, \ourm needs the information spread across tables. For example in Figure ~\ref{fig:intro}, a actor-genre and director-genre helps model the genre-preferences for both entities and thus might help a director find a plausible actor for his movie. 
However, learning entity representations over all the tables bears a significant computation cost as computing (either explicitly or implicitly) all pairs of fk-pk joins between tables --also termed as full \textit{denormalization}-- of the database. Secondly, a denormalized-table corrupts data quality as it can potentially combine unrelated attributes as well as drastically increases the table size\sbcomment{I am not sure of about this statement}. To avoid these bottlenecks, we model the table joins as a \textit{next-sentence} prediction problem. Specifically, consider two tables $\bs{T}_\alpha$ and $\bs{T}_\beta$ that can be joined via the primary-key column $\bs{T}^A_{\alpha}$, and the corresponding \textit{foreign-key} column $\bs{T}^{\hat{A}}_\beta$. If a row with entity $\tau_i \in \bs{T}^A_\alpha$ is linked to the row with $\tau_j \in \bs{T}^{\hat{A}}_\beta$ then the latter is considered as the \textit{next sentence} for the current row.  
We utilize the output embedding at the $\texttt{[CLS]}$ token to estimate the probability of a sentence being the \textit{next} sentence, $v_{\tau_i, \tau_j}$ and use a negatively sampled row $\tau'_j \in \bs{T}^{\hat{A}}_\beta$. The loss for NSP is calculated via:
\begin{equation}
\cm{L}_{nsp} = -\sum_{i=1}^{|\bs{T}_\alpha|} \sum_{\tau_i \in \bs{T}^A_\alpha} \left[ \log \left ( \sigma(v_{\tau_i, \tau_j}) \right) + \log \left (1 - \sigma(v_{\tau_i, \tau'_j}) \right) \right],
\end{equation}
where $\tau_j, \tau'_j \in \bs{T}^{\hat{A}}_\beta$ and $\sigma$ denote the \textit{true}, the sampled negative sentences and \textit{sigmoid} function respectively.
Finally, we combine this next sentence prediction task with the mask language model learning of the predefined table encoder.

\subsection{Optimization}
Training our model for table MLM enables us to learn relationships between attributes whereas the NSP task helps to aggregate inter table information i.e. join between the tables. By introducing the next sentence prediction task, we develop the first table auto-completion approach that avoids the bottlenecks of a denormalised table. 
The parameters for \ourm-A  are optimized by minimizing the net loss which is the sum of the corresponding MLM and NSP cross entropy losses:
\begin{equation}
\cm{L}_{A} = \cm{L}_{mlm} + \cm{L}_{nsp},
\end{equation}
However, joining different tables at each training iteration would lead to a information loss over multiple iterations. Therefore, we our training strategy consists of a two-step approach, with the first step as general \textit{pre-training} step which is followed by task-specific \textit{fine-tuning}. We describe them here:

\paragraph{Pre-training:} To understand the relationships between tables and attributes during pre-training, we perform a task-independent learning of all entities in the database. Specifically, we construct pairwise sentences between tables using the primary-key joins in the relational schema and train the model parameters using the standard MLM-based optimization. 

\paragraph{Fine-Tuning:} The pre-training step is followed by task-specific \textit{fine-tuning} of the model. More so, for our task of table auto-completion we jointly fine-tune the \ourm-A model to predict the \textit{masked} entity for each row and simultaneously the join-based next sentence by optimizing the net loss $\cm{L}_{A}$.

\subsection{\ourm-J}
An integral function for a database is to \textit{join} multiple tables i.e, link rows corresponding to two entities in different tables based on an existing \textit{primary-foreign} key relation. Computing a join is a trivial task in RDBMS when exact matches exist between different attributes. However in a neural setting, the entities are represented via embedding vectors that capture the inter- and intra-table relationships. Thus identifying identical entities is an impractical task as these embedding can vary with different tabular structures. This is further exacerbated by the presence of missing entries in the relational tables.  If the database embeddings are not consistent with the table-joins then the critical information of database schema will inherently be lost. Unfortunately, previous approaches ~\cite{EmbDi}, \cite{Tab2Vec} overlook this issue and focus on on the solitary goal of learning representations for entities rather than the database as a whole. Our proposed variant \ourm-J ascertains if we can combine embeddings from the models learn on table and effectively extend it to learn embedding for the entire relational database. 
In this variant, we run independent masked-language models on each table in the database. Specifically for tables with related columns $\bs{T}^A_\alpha$ and $\bs{T}^{\hat{A}}_\beta$ we learn entity embeddings as $\bs{P}_{\alpha} = \texttt{MLM}(\bs{T}_\alpha)$ and $\bs{P}_{\beta} = \texttt{MLM}(\bs{T}_\beta)$ respectively. Here, $\texttt{MLM} (\bullet)$ denotes function MLM based transformer model. Later, to capture the relationships between tables, we optimize the NSP loss between between the tables that have a relational join between them. 
\begin{equation}
\cm{L}_J = \texttt{NSP} (\bs{P}_{\alpha}, \bs{P}_{\beta}, \bs{T}^A_\alpha, \bs{T}^{\hat{A}}_\beta)
\end{equation}
where $\cm{L}_J, \texttt{NSP}$ denote the net loss for \ourm-J and the transformer learned via next sentence prediction.
Thus a major distinction between \ourm-A and \ourm-J is that the former trains on a join of tables whereas in latter, we independently pre-train on different tables and \textit{fine-tune} on a table join.
\section{Experiments} \label{sec:expt}
In this section we assess the quality of the embeddings learned by both \ourm models by putting them at test for two tasks -- cell autocompletion and join prediction. With our experiments we aim to answer the following research questions:
\begin{description}
\item[\textbf{RQ1}] Can \ourm-A outperform state-of-the art baselines for Table Autocompletion?
\item[\textbf{RQ2}] How well can \ourm-J join two tables in the database?
\item[\textbf{RQ3}] What is the scalability of both approaches? 
\end{description}
All the codes for our proposed models are implemented using Pytorch \footnote{\url{https://pytorch.org}}. We train our model as well as the baselines using an NVIDIA GTX 1080 Ti GPU and a machine with Intel(R) 64 CPUs and 337 GB memory. 
\subsection{Setup} 
\paragraph{Dataset:} We used the following two relational databases for evaluating the performance of \ourm:
\begin{itemize}
    \item \textbf{IMDB \footnote{\url{https://relational.fit.cvut.cz/dataset/IMDb}}}: A database that serves as an encyclopedia of movies, web-series, documentries and etc. It comprises of 7 tables with 21 columns overall and aggregate $5,694,919$ tuples.
    
    \item \textbf{MIMIC \footnote{\url{https://mimic.physionet.org/}}}: A real database capturing information about the anonymous patients admitted to Beth Israel Deaconess medical center. Every aspect of a patient's visit, like Diagnostics, is recorded. For our experiments, we chose 6 tables with densely connected 12 fk-pk joins.  The tables contain a total of $3,583,700$ rows and $42$ columns.
%
%
\end{itemize}
\paragraph{Baseline Models:} For table autocompletion task, we compared our model with the following state-of-the-art database embedding approaches:
\begin{itemize}
\item \textbf{Table2Vec \cite{bordwaker}}: This model treats each row of a table as a sentence and trains a word2vec model on the corpus of such sentences. We treat a single sentence as the context, and consequently, the context window size is same as that of the sentence. We trained the model for entity embeddings of size $300$.
    
\item \textbf{EmbDi \cite{EmbDi}}: The state-of-the-art approach that creates a tripartite graph from data where each entity is connected to its row id and attribute id node in the graph. Then they create sentences from this graph by running multiple random walks over each token node and apply a standard NLP-based embedding approach i.e., word2vec over this sentence corpus, to generate embeddings. For sentence construction, we perform $1,000$ random walks for each entity of length $60$ over the tripartite graph. 
\end{itemize}

\paragraph{Tasks:}
For autocompletion or missing value prediction task, we consider a situation where two tables, $\bs{T}_\alpha$ and $\bs{T}_\beta$, are joined with primary-foreign key join between $\tau_i \in \bs{T}_\alpha$ and $\tau_j \in \bs{T}_\beta$. Then the task is to predict a ranked list of candidate entities for the missing cell in new table.  For IMDB dataset we consider each director and randomly sample 70\% of her movies in the training set, 15\% in the validation set, and 15\% of movies in the test set. Then, for every row containing (director, movie) pair in the validation/test set, we break the row into two sequences. The first sequence consists of movie-related information (i.e., join of movies, actors, roles, movies\_genres table), and the second sequence consists of director related information (i.e., join of directors and directors\_genres table). We then aim to predict the correct director for each movie from the entire list of directors in our dataset sing just the actor and director information.
\noindent For MIMIC III, we aim to predict the \texttt{DRGCODE} (diagnosis-related groups) for each patient in the dataset. Similar to IMDB, for each entity in the \texttt{DRGCODE} column we divide subjects in the training, validation and the test set. Then we break each row (\texttt{Patient}, \texttt{DRGCODE}) in the validation and test set, we break the row into two sequences: (1) patient-related information (i.e., from the join of \texttt{ADMISSIONS}, \texttt{PATIENTS}, \texttt{CPTEVENTS}, \texttt{PROCEDURES\_ICD}, and \texttt{DIAGNOSES\_ICD} tables), (2) Diagnosis group related information from \texttt{DRGCODE} table.

\noindent For join-prediction, we have focused on joins that are dictated by primary key and foreign key pairs in the database and compute the average join prediction performance over all such joins. For IMDB, we worked with \texttt{Movies} and \texttt{Movies-Directors} relation that are joined over \texttt{MovieId} column attribute. \\
\begin{figure}[t]
\centering
\begin{subfigure}[b]{0.32\columnwidth}
\includegraphics[scale = 0.45]{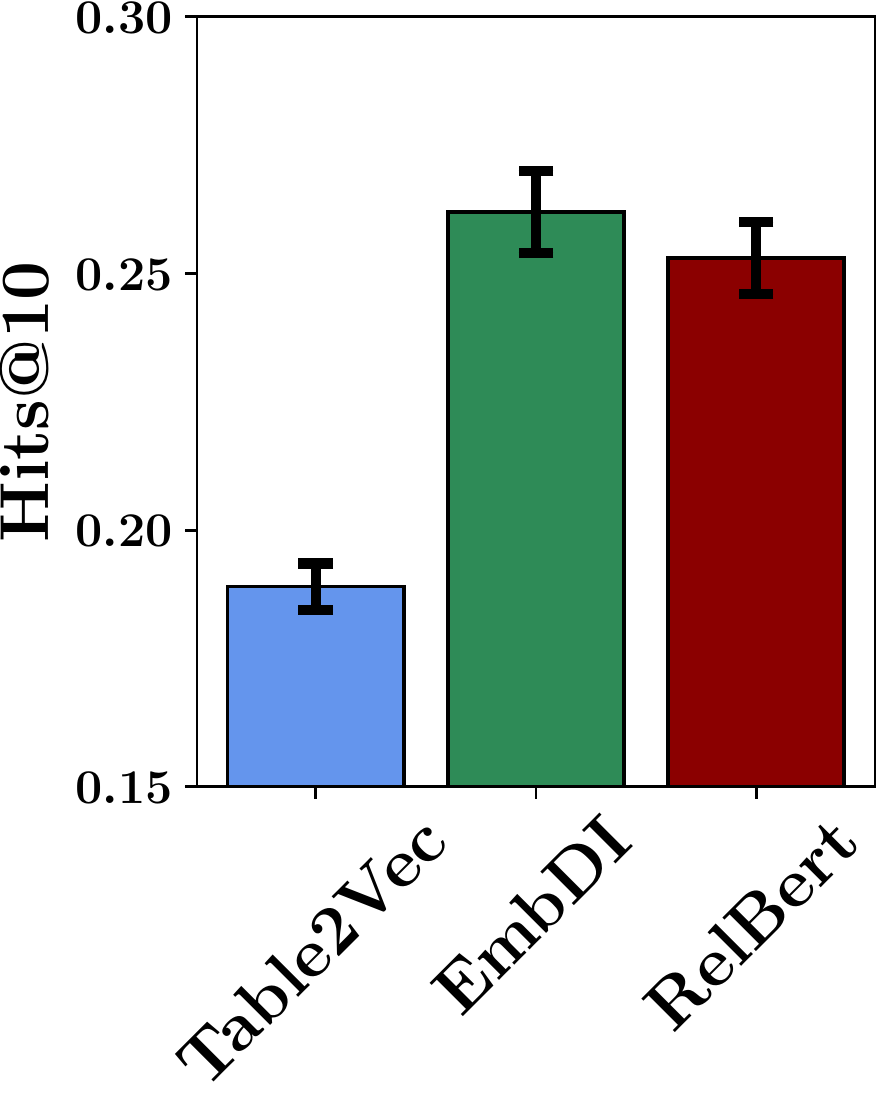}
\caption{Hits@10 $\uparrow$}
\end{subfigure}
\begin{subfigure}[b]{0.32\columnwidth}
\includegraphics[scale = 0.45]{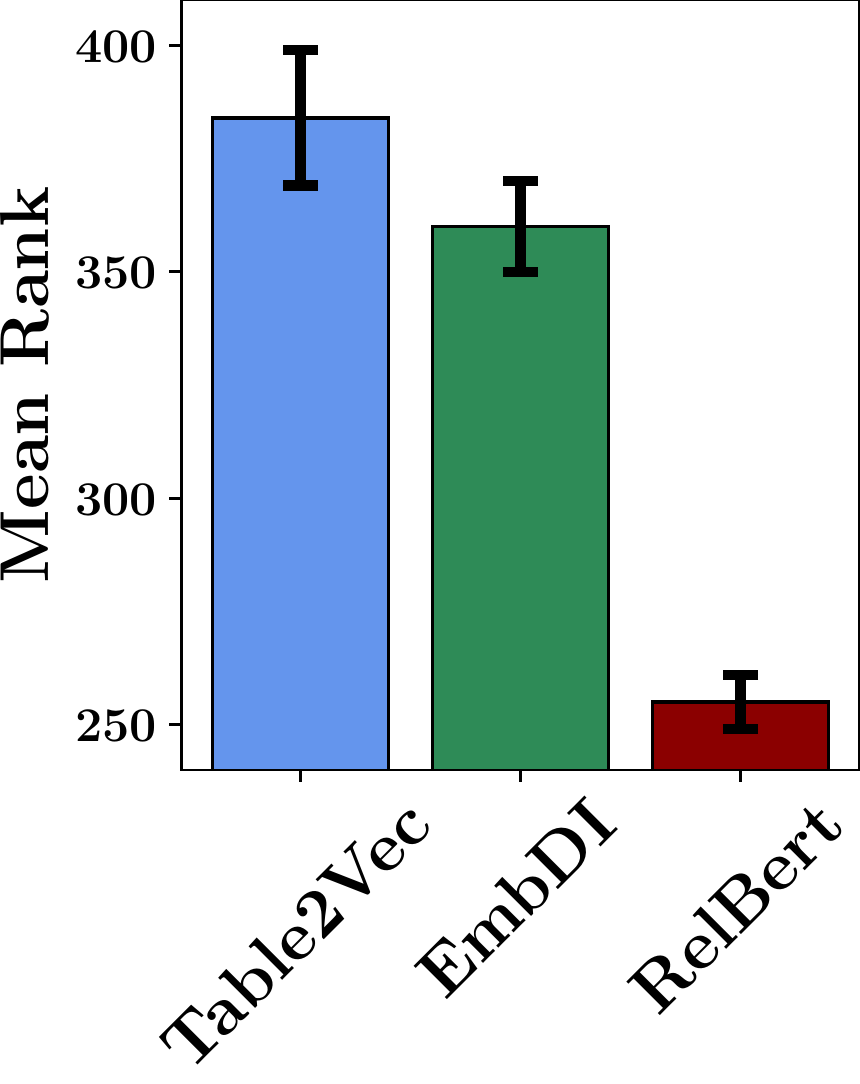}
\caption{MR $\downarrow$}
\end{subfigure}
\begin{subfigure}[b]{0.32\columnwidth}
\includegraphics[scale = 0.45]{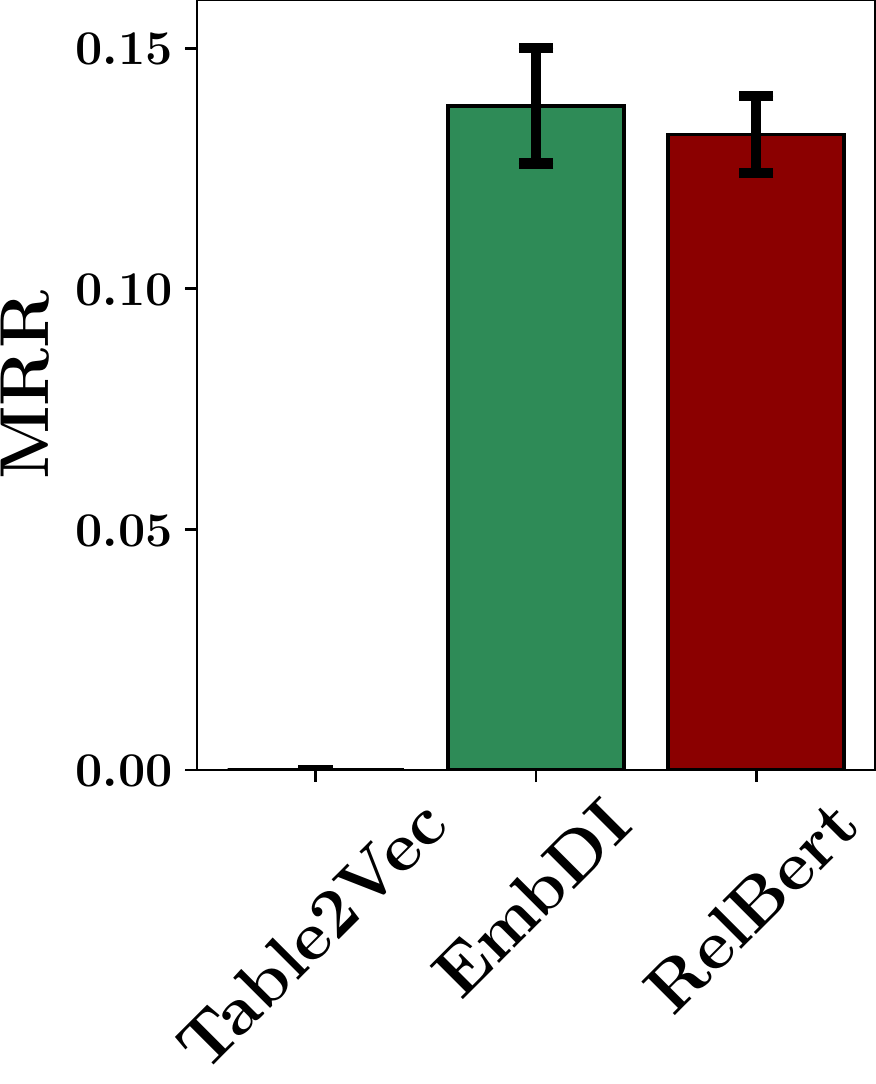}
\caption{MRR $\uparrow$}
\end{subfigure}
\caption{\label{fig:main_imdb} Results for the Auto Completion of Table Cells on test set for IMDB dataset. Upper (lower) arrow indicates that higher the value, better (worse) the performance.}
\end{figure}

\begin{figure}[t]
\centering
\begin{subfigure}[b]{0.32\columnwidth}
\includegraphics[scale = 0.45]{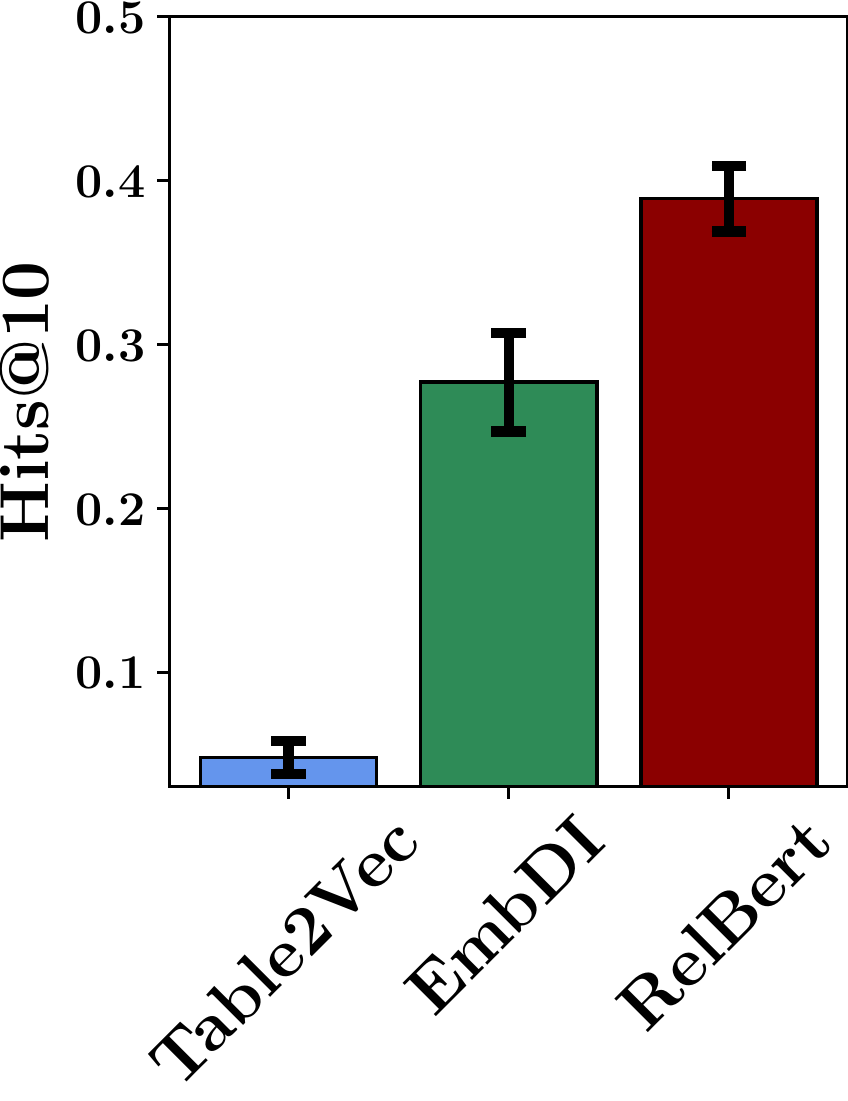}
\caption{Hits@10 $\uparrow$}
\end{subfigure}
\begin{subfigure}[b]{0.32\columnwidth}
\includegraphics[scale = 0.45]{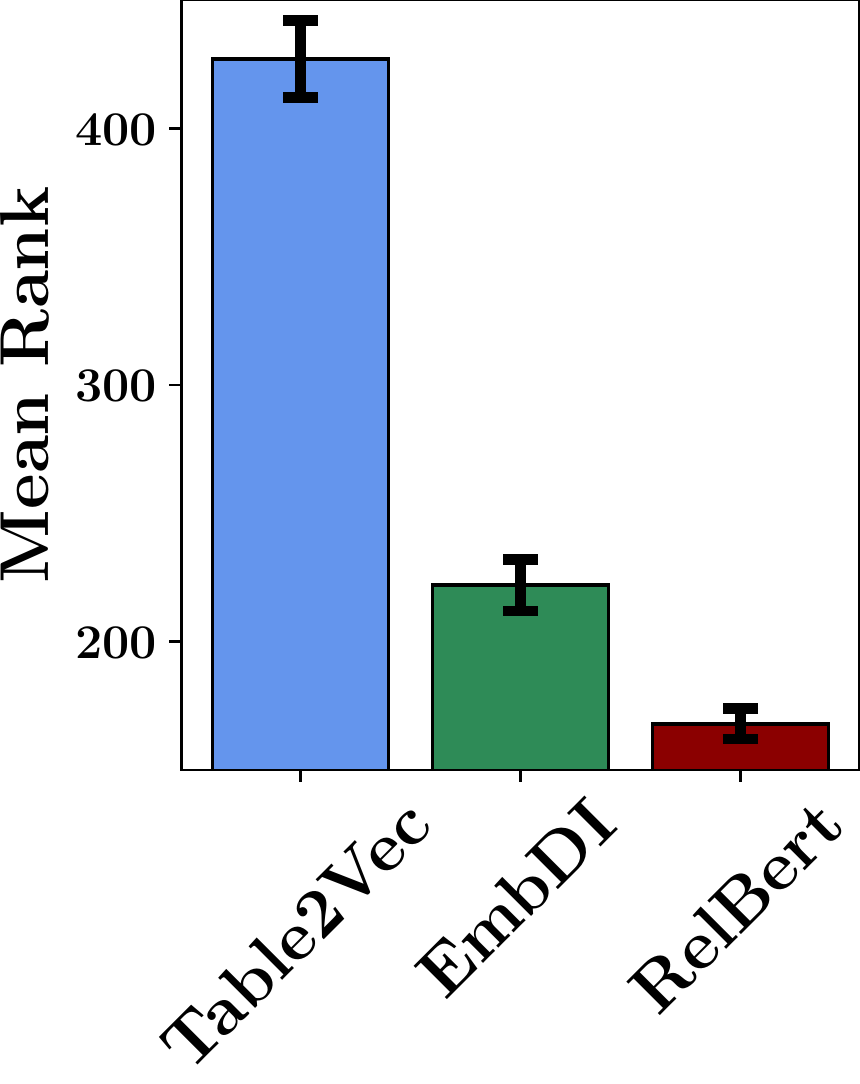}
\caption{MR $\downarrow$}
\end{subfigure}
\begin{subfigure}[b]{0.32\columnwidth}
\includegraphics[scale = 0.45]{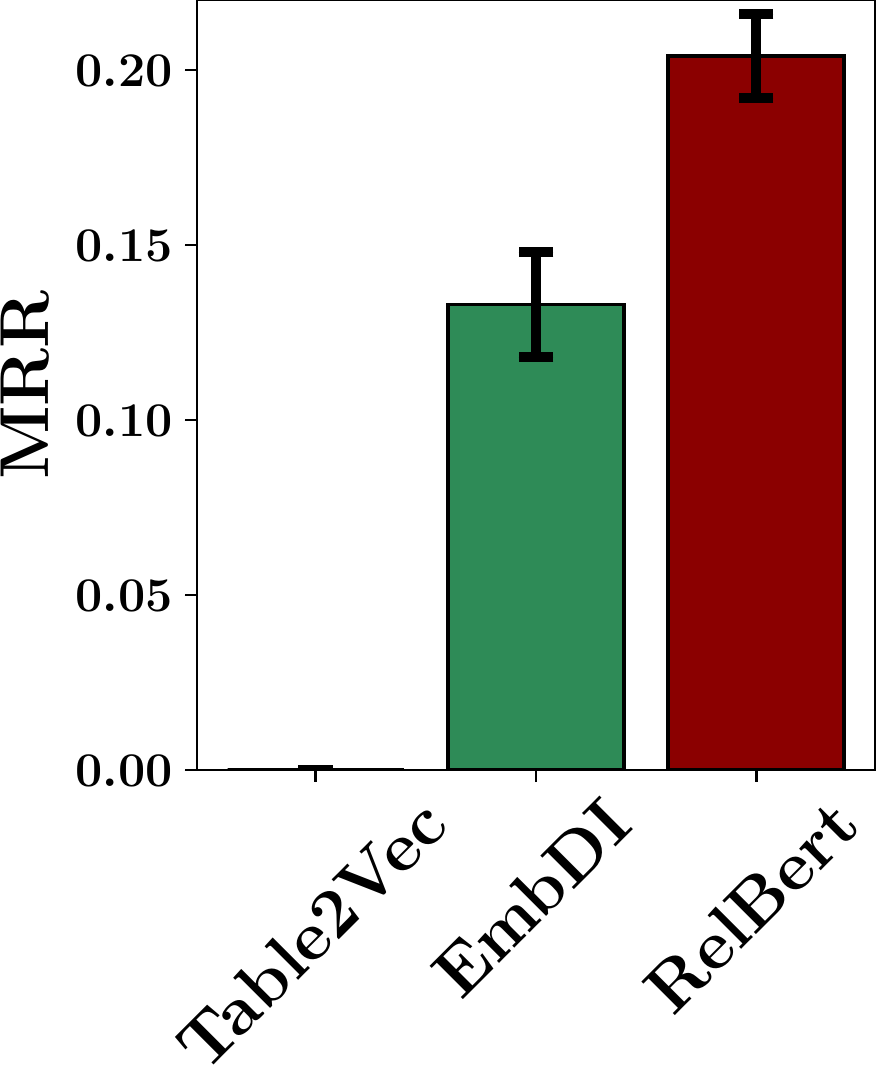}
\caption{MRR $\uparrow$}
\end{subfigure}
\caption{\label{fig:main_mimic} Results for the Auto Completion of Table Cells on test set for MIMIC dataset. Upper (lower) arrow indicates that higher the value, better (worse) the performance.}
\end{figure}

\paragraph{Metrics:} We compare models across three metrics: Hits@10,  \emph{mean rank} (MR) and \emph{Mean Reciprocal Rank} (MRR). 

\subsection{Auto-completion (RQ1)}
The results for the autocompletion for IMDB and MIMIC are given in Figure \ref{fig:main_imdb} and Figure \ref{fig:main_mimic} respectively. From the results, we make the following observations:
\begin{itemize}
    \item \ourm-A consistently outperforms Table2vec~\cite{bordwaker} for both datasets. This can be mainly attributed to the fact that column entities have a dissimilar importance for predicting the missing entities. Therefore, a differential column weighting mechanism of \ourm-A helps to better aggregate the tabular information, that is overlooked by Table2Vec.
    
    \item The recently proposed state-of-the-art model EmbDi~\cite{EmbDi} also outperforms Table2Vec. For IMDB, we observe that \ourm perfroms comparable to EMbDi, however it easily outperforms all baselines for the Mimic dataset. We note that in IMDB, the entities among different tables occur more frequently than in Mimic. Understandably, an actor and director would have much more information that a patient's treatment records. We note that in such cases EmbDi is more suitable as the random-walk based model can learn efficient embeddings if it frequently visits a node in a graph. However, this benefit comes at the expense of a large computation cost that we explore later. 
\end{itemize}

\subsection{Qualitative Analysis}
\begin{figure}[t]
  \centering
    \includegraphics[width=\linewidth]{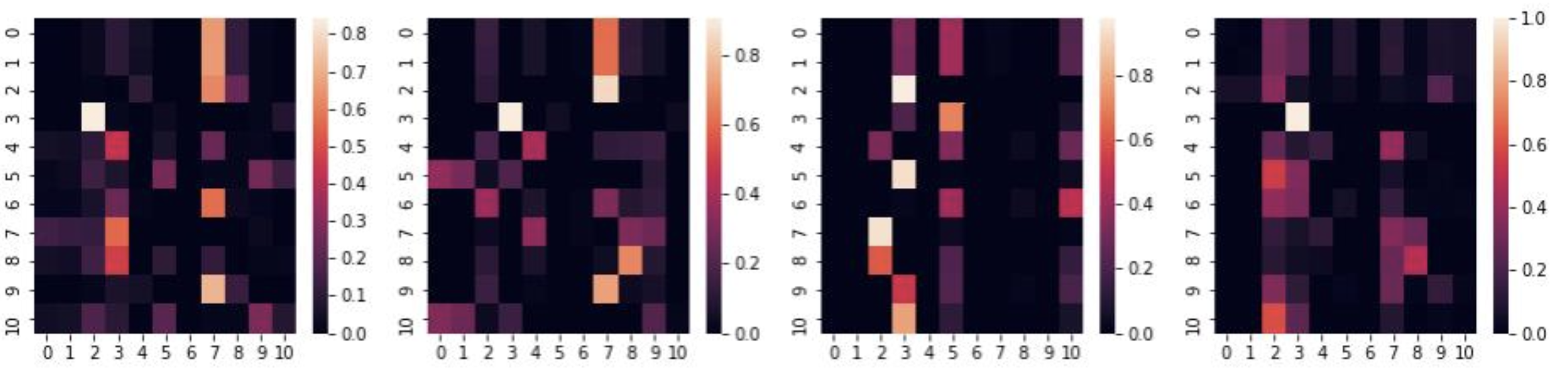}
  \caption{Attention Heatmaps of the first layer of encoder layer for 4 different attention heads for IMDB dataset.}
  \label{fig:hmap}
\end{figure}
To analyze the contribution of the self-attention model in capturing the column-semantics, we plot the attention heatmaps for the IMDB dataset in Figure \ref{fig:hmap}. These maps correspond to the first table-encoder layer of our model. Labels [1$\rightarrow$12] specify the column ordering [\texttt{movie\_id}, \texttt{movies\_name}, \texttt{movies\_year}, \texttt{movies\_rank}, \texttt{actors\_last\_name}, \texttt{movies\_genre}, \texttt{actor\_id}, \texttt{role}, \texttt{actors\_first\_name}, \texttt{directors\_id}, \texttt{directors\_genre}] with row signifying the \textit{query} and the column, the \textit{key} in attention layer. From the figure, we note that transformer aggregates the contribution from all columns and do not simply prefer columns that co-exist together in table. We also note that different attention heads concentrate on different aspects of database. The first and second attention heads are more focused on actor and role information, the third is centered around movie- and director-genres fields and the last attention head is captures the movie-rank and movie-year. We exclude the plots for Mimic datasets as they also demonstrated similar trends.

\subsection{Join Prediction (RQ2)}
In this section we evaluate the join-prediction performance of \ourm. This is a novel contribution to the neural-database literature and to the best of our knowledge none of the  baseline models directly address this problem. Therefore we can only compare with the \textit{gold-standard} for join-prediction, i.e, a standard database join. DUe to brevity purposes, we report results for the IMDB dataset. To evaluate the join-prediction model, we consider \texttt{Movies} and \texttt{Movie-Director} table and train the encoder using  independent MLM based optimization on both the tables and then fine-tune using the NSP model to predict all the joins between these two tables. More details are given in Section 3.4. We further experiment with the no of negatively sampled next sentences for fine-tuning the NSP.
\begin{table} [h]
\centering
\begin{tabular}{@{}l l l l @{}} 
 \toprule
 Sampling Count & Hits@10 & MR & MRR \\
 \midrule
 \# Samples $=1$ & 0.328 & 510.23 & 0.140 \\ 
 \# Samples $=5$ & 0.730 & 476.29 & 0.507 \\
 \# Samples $=10$ & 0.801 & 284.25 & 0.656 \\
\bottomrule
\end{tabular}
\caption{Results for \ourm-J in Join-Prediction between Actor and Director tables IMDB dataset.}
\label{tab:join}
\end{table}
From the results in Table \ref{tab:join}, we note that computing the \textit{exact} join based on neural embedding is not a trivial task. However, the prediction accuracy of 80\% at Hits@10 further reinforces our claim that \ourm-J is able to effectively capture the relational structure and predict the joins between the tables. We also note that by increasing the negative samples for NSP fine-tuning, the prediction accuracy also increases. We consider the exact parameter tuning as a future work of our paper.

\subsection{Scalability (RQ3)}
\begin{figure}[h]
\centering
\includegraphics[scale = 0.45]{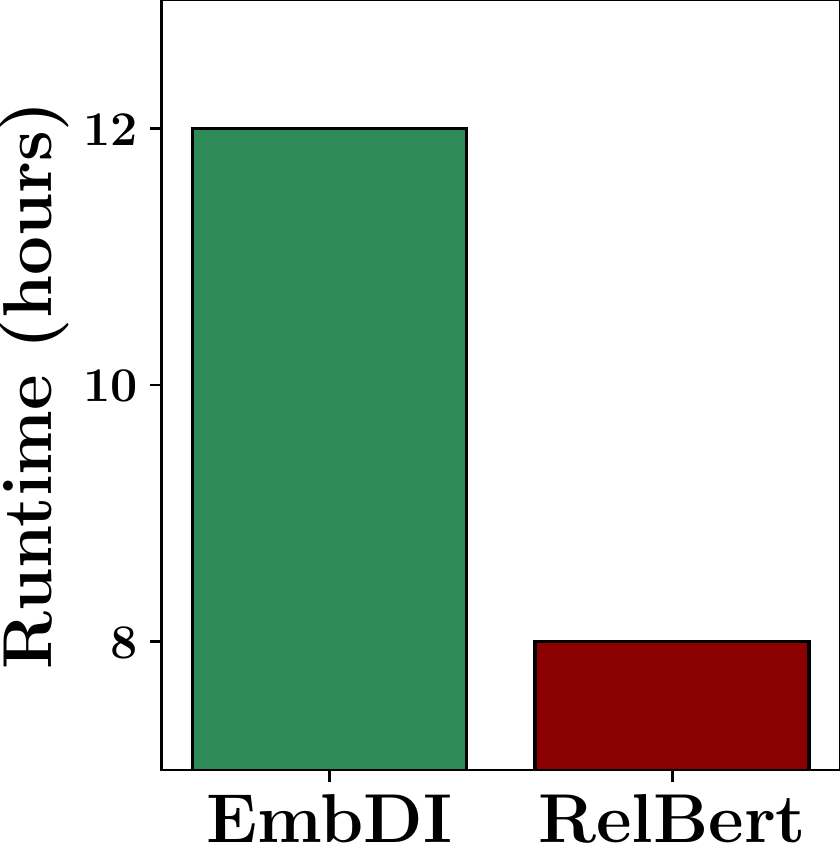}
\caption{Comparison of Running Time of \ourm-A and EmbDI over IMDB}\label{fig:time}
\end{figure}

Finally we report on the computational efficiency of \ourm and its variants to verify its applicability in real-world settings. As mentioned in RQ1, the random walks in EmbDI are expensive to compute if entities among different tables occur more frequently. We compare the run-times of \ourm-A and EmbDI for the IMDB dataset in Figure ~\ref{fig:time} and note that \ourm is almost 50\% faster than EmbDI and still attains similar prediction performance. The run-times for training the \ourm-J variant is similar to \ourm-A due to their mostly similar architectures. The fine-tuning process for \ourm only requires $26$ minutes. Therefore it shows the ability of \ourm to learn any task quickly after an initial pre-training rather than re-training any other model. Thus the training as well as fine-tuning times for \ourm its variants are feasible for real-world deployment.

\section{Conclusion}
As we can see from all the result tables, there are still vast improvements that need to be made to learn useful entity embeddings that can improve over downstream tasks like solving semantic queries and auto-completion of data cells in relational databases. There are a lot of exciting directions that we want to look at in our future work. We would first want to try vertical attention transformers to capture intra-attribute relationship. We are also very curious to investigate learning query biased embeddings, i.e., tune the embeddings of databases based on the SQL query run on the database. We aim to examine the application of learned representation for entries in the database for the semantic join of tables, particularly query-sensitive semantic join of tables in the relational database. 

In this work, we propose the first approach that can learn the embeddings for relational databases and is scalable for real-world datasets. We also see that our approach achieves comparable or better results than learning embeddings on complete denormalised table and other existing baselines.
\bibliographystyle{unsrt}  
\bibliography{ref}

\clearpage

\section*{\ourm: Supplementary Material}

\subsection*{Dataset Details}
\begin{figure}[h]
    \centering
    \begin{minipage}{0.4\textwidth}
        \centering
        \includegraphics[width=\textwidth]{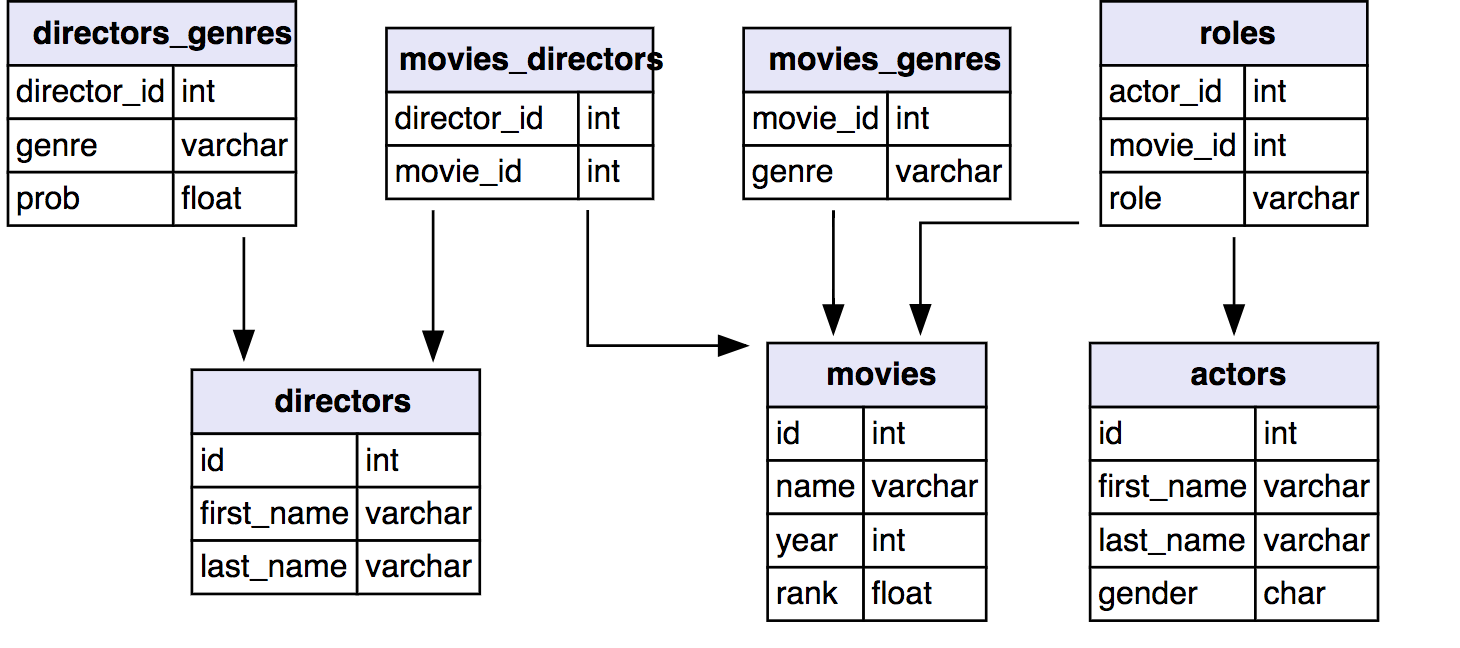}
    \end{minipage}
    \caption{The Schematic Structure of the IMDB dataset}
    \label{fig:IMDB}
\end{figure}

\paragraph{IMDB}: The schema of the database is shown in Figure~\ref{fig:IMDB}. The connections between tables indicate the PK-FK pairs of the database, and thus makes the connected tables join-compatible. Recall that while training, we map entities from different columns to different latent space, and while testing, masked tokens are associated with  columns and the task is to find the most probable entity in the respective latent space. In Table \ref{tab:entity_count}, we report the number of unique entities that belong to each column of all the tables. We clean the data by removing all the directors and actors who have worked in less that $5$ movies. We also removed movies with movie rank field as \texttt{NONE}.

\begin{table}[h]
\centering
 \begin{tabular}{@{}l l  @{}} 
 \toprule
 Column Name & \#Entity \\
 \midrule
 director\_id & 3236 \\
 movie\_id & 32119 \\
 actor\_id & 57221 \\
 role & 287001 \\
 actor\_first\_name & 14361 \\
 actor\_last\_name & 30978 \\
 movie\_genre & 20 \\
 movie\_name & 31124 \\
 movie\_year & 108 \\
 movie\_rank & 9 \\
 director\_genre & 20 \\
 \bottomrule
\end{tabular}
\caption{Entity count for each column space in dataset}
\label{tab:entity_count}
\end{table}

\begin{figure}[h]
    \centering
    \begin{minipage}{0.4\textwidth}
        \centering
        \includegraphics[width=\textwidth]{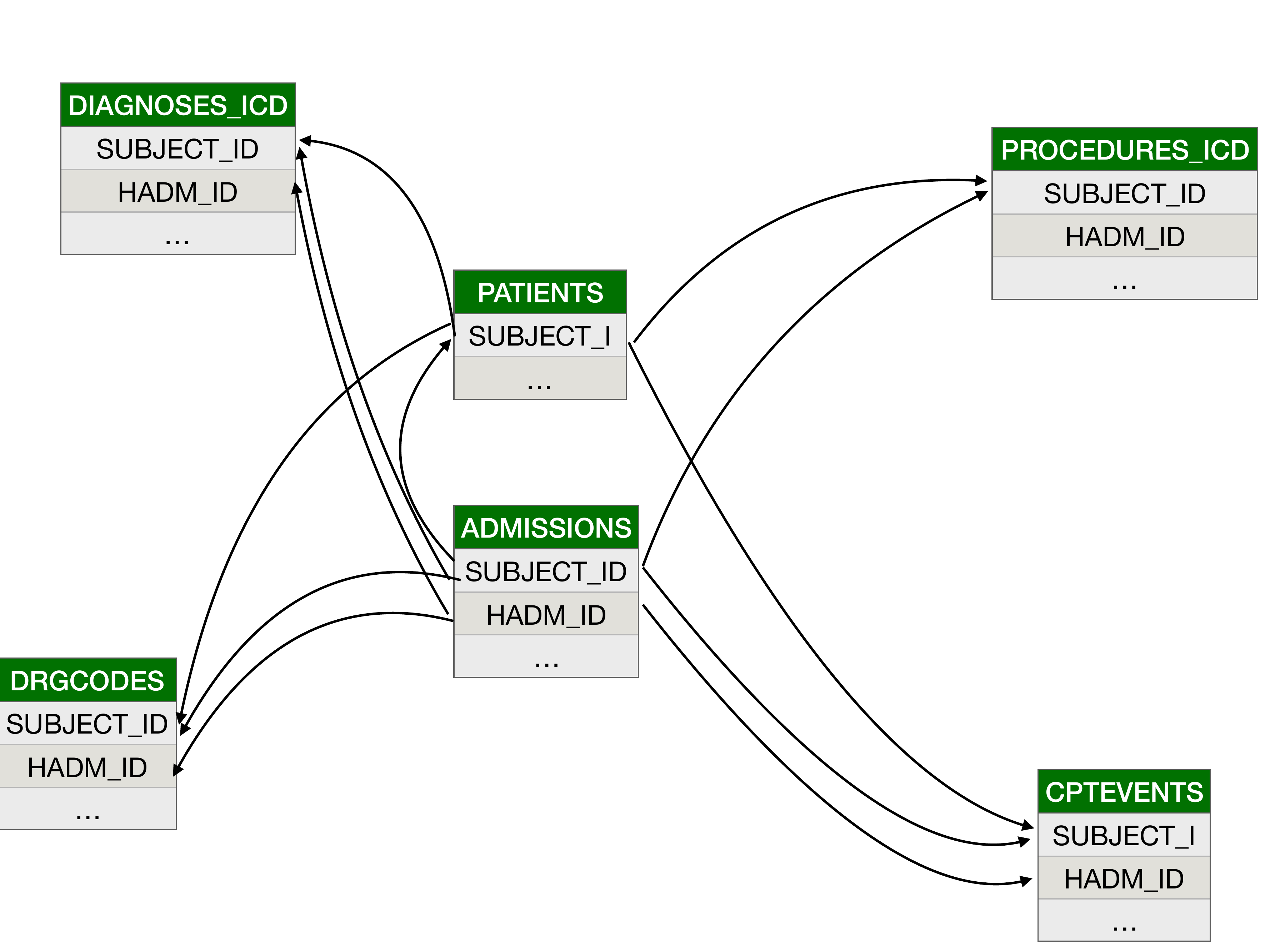}
    \end{minipage}
    \caption{The Schematic Structure of the MIMIC dataset.}
    \label{fig:MIMIC}
\end{figure}

\paragraph{MIMIC} The schema and the data statistics of tables are given in Figure~\ref{fig:MIMIC} and Table~\ref{tab:mimic} respectively. Here, we clean data by removing patients who have a history of more than one hospitalization. We also remove patients with more than $10$ diagnoses since it could be too general to make any specific predictions.
\begin{table}[h]
\centering
 \begin{tabular}{@{}l l l @{}} 
 \toprule
 Table Name & Col & Row \\
 \midrule
 Admissions & 18 & 12,399\\
 Cptevents	& 11 & 89,208\\
 Diagnoses\_ICD	& 4 & 90,379\\
 Drgcodes &	7 & 24,784\\
 Patients &	7 & 12,399\\
 Procedures\_ICD	& 4 & 54,531\\
 \bottomrule
\end{tabular}
\caption{Description of subset of tables in the filtered MIMIC dataset that are used for our experiments }
\label{tab:mimic}
\end{table}

\subsection*{Parameter Settings}
For the encoder based architectures, we use Embedding dimension 300, followed by 4 layers of transformer encoder for MIMIC and 6 layered transformers for IMDB. Each layer has multi-head attention with 4 heads and Feed Forward Network with hidden dimension 1200. For the joint NSP and MLM pre-training and fine-tuning based  models we use Embedding dimension 300, followed by 4 layers of transformer encoder for both MIMIC and IMDB. Other configurations remain same as before.
\end{document}